%% file: main.tex
\documentclass[a4paper, 10pt, conference]{IEEEtran}
\IEEEoverridecommandlockouts
\usepackage{cite}
\usepackage{amsmath,amssymb,amsfonts}
\usepackage{algorithmic}
\usepackage{graphicx}
\usepackage{textcomp}
\usepackage{todonotes}
\usepackage{lipsum}
\usepackage{tabularx}
\usepackage{booktabs}
\usepackage{multirow}
\usepackage{slantsc}
\usepackage{xltabular}
\usepackage{tabularray}
\usepackage{float}
\usepackage{multicol}
\usepackage{stfloats}
\usetikzlibrary{shapes,snakes}
\usetikzlibrary{shapes.geometric}
\usepackage{svg}
\usepackage[section]{placeins}
\usepackage[caption=false]{subfig}
\usepackage[nolist]{acronym}
\usepackage{hyperref}
\usepackage{xcolor}

\include{content/definitions}

\begin{document}
	\input{content/title}
	\input{content/acronyms}
	\input{content/0_abstract}
	\input{content/1_introduction}
	\input{content/2_system}
	\input{content/3_concept}
	\input{content/4_conclusion}
	\input{content/acknowledgments}
	\bibliographystyle{IEEEtranBST/IEEEtran}
	\bibliography{IEEEtranBST/IEEEabrv,references}
\end{document}

%% file: content/definitions.tex
\definecolor{whitesmoke}{rgb}{0.96, 0.96, 0.96}

\usetikzlibrary{arrows,chains,positioning,fit}
\newcommand{\krug}[1]{\tikz[baseline=(char.base)]{\node[shape=circle,draw,minimum size=4mm, inner sep=1pt, semithick](char){#1}}}
\newcommand{\kvadrat}[1]{\tikz[baseline=(char.base)]{\node[shape=rectangle,draw,minimum size=4mm, inner sep=1pt, semithick](char){#1}}}
\newcommand{\peterokut}[1]{\tikz[baseline=(char.base)]{\node[shape=regular polygon,regular polygon sides=5,draw,minimum size=4mm, inner sep=1pt, semithick](char){#1}}}

\def\BibTeX{{\rm B\kern-.05em{\sc i\kern-.025em b}\kern-.08em
		T\kern-.1667em\lower.7ex\hbox{E}\kern-.125emX}}

\def\rot#1{\rotatebox{90}{#1}}

\newcommand\fauxsc[1]{\fauxschelper#1 \relax\relax}
\def\fauxschelper#1 #2\relax{%
	\fauxschelphelp#1\relax\relax%
	\if\relax#2\relax\else\ \fauxschelper#2\relax\fi%
}
\def\Hscale{.85}\def\Vscale{.74}\def\Cscale{1.12}
\def\fauxschelphelp#1#2\relax{%
	\ifnum`#1>``\ifnum`#1<`\{\scalebox{\Hscale}[\Vscale]{\uppercase{#1}}\else%
	\scalebox{\Cscale}[1]{#1}\fi\else\scalebox{\Cscale}[1]{#1}\fi%
	\ifx\relax#2\relax\else\fauxschelphelp#2\relax\fi}

%% file: content/title.tex
\title{Driverless road-marking Machines: Ma(r)king the Way towards the Future of Mobility\\
}

\author{\IEEEauthorblockN{Domagoj Majstorovi\'c}
	\IEEEauthorblockA{\textit{Institute of Automotive Technology} \\
		\textit{Technical University of Munich}\\
		Garching bei M{\"u}nchen, Germany \\
		domagoj.majstorovic@tum.de}
	\and
	\IEEEauthorblockN{Frank Diermeyer}
	\IEEEauthorblockA{\textit{Institute of Automotive Technology} \\
		\textit{Technical University of Munich}\\
		Garching bei M{\"u}nchen, Germany \\
		diermeyer@tum.de}
}

\maketitle

%% file: content/acronyms.tex
\begin{acronym}
	\acro{rnd}[R\&D]{research and development}
	\acro{cots}[COTS]{commercial off-the-shelf}
	\acro{ad}[AD]{Automated Driving}
	\acro{ar}[AR]{Augmented Reality}
	\acro{qa}[QA]{Quality Assurance}
	\acro{ros}[ROS]{Robot Operating System}
	\acro{cnn}[CNN]{Convolutional Neural Network}
\end{acronym}

%% file: content/0_abstract.tex
\begin{abstract}
Driverless road maintenance could potentially be highly beneficial to all its stakeholders, with the key goals being increased safety for all road participants, more efficient traffic management, and reduced road maintenance costs such that the standard of the road infrastructure is sufficient for it to be used in \ac{ad}.
This paper addresses how the current state of the technology could be expanded to reach those goals.
Within the project `System for Teleoperated Road-marking' (SToRM), using the road-marking machine as the system, different operation modes based on teleoperation were discussed and developed.
Furthermore, a functional system overview considering both hardware and software elements was experimentally validated with an actual road-marking machine and should serve as a baseline for future efforts in this and similar areas.
\end{abstract}


%% file: content/1_introduction.tex
\section{Introduction}

In most countries, roads are the major type of transport infrastructure for both freight and passengers, with road networks facilitating transport services and making travel more time/energy efficient while reducing the overall costs of trade. They are built in all kinds of land topography, diverse geographical locations and are affected by weather and different natural occurrences such as flooding and landslides. Additionally, their condition deteriorates over time from wear and tear caused by traffic and, often, neglect. To achieve the continuity of road transport benefits, road infrastructure needs to be continuously maintained in a good condition. This maintenance includes different activities such as road paving, lane painting, paint and bead striping, asphalt cleaning, patch filling, sealing of cracks, grass mowing, snow sweeping, sign (re-)installation, etc. \cite{Huang2019-zn} While the requirements for road maintenance vary from place to place, it is always a continuous process that needs to meet specific requirements to ensure the safety of all road participants.

\subsection{Motivation}
The performance of road maintenance is often dangerous for both traffic and maintenance crews. Studies in recent years have shown that the construction zones in and around highway areas are associated with an increase in death and injuries [\citen{Wang1996-sn}, \citen{Yang2015-im}]. Apart from safety, efficiency is another keyword for highway construction zones. Speed  limits and full lane closures imposed by infrastructure authorities are necessary to protect both road workers and road users, leading to congestion and delays. \cite{ADB_Independent_Evaluation_Department2013-wm}

Given  these  negative  traffic  impacts, the idea has emerged of replacing road workers with driverless  maintenance  vehicles.
This idea has become even more relevant in recent years given the increase in \ac{rnd} activities in the field of teleoperation and autonomous driving of passenger vehicles.
Huang \cite{Huang2019-zn} has conducted a feasibility study of driverless maintenance in highway construction zones in the Netherlands and showed that employing such vehicles as part of the road maintenance chain would have a multitude of positive side effects for all stakeholders.
Table \ref{tab:stakeholders} lists some of the possible benefits for the major shareholders.
Other shareholders, such as vehicle authorities, research institutions, insurance companies, vehicle manufacturers, etc., would also benefit from utilizing these technologies.

\input{tables/1_stakeholders.tex}

\subsection{State of Technology}
\label{subsec:SoT}
Over the course of the last two decades, there has been a lot of \ac{rnd} aimed at bringing some level of automation to different road maintenance tasks with the main goal of improving safety and work efficiency while reducing operating costs.
As long ago as 2007, an autonomous weeder platform, previously used for agricultural activities in Netherlands, was introduced as a system for a case study assessing the potential for autonomous road maintenance \cite{Bakker2007}.
The study showed that the technology at the time was not mature enough to achieve driverless sweeping/mowing, and further \ac{rnd} was needed.
In 2013 Japan Automobile Research Institute and Naka Nippon Expressway first attempted to introduce automated driving technology into road maintenance activities by use of a vehicle for automated cleaning of tunnel infrastructure.
This resulted in a significant reduction in downtime \cite{jari}.
In~2015, Stolte et al. \cite{Stolte2015-kh} introduced a prototype of a driverless protection vehicle that was supposed to support road maintenance activities and autonomously follow the working vehicle at a close distance and low speed of approximately 10 km/h.
They introduced various operation modes that served as a basis for various hazard assessments in later research (HAZOP, FMEA, etc.) [\citen{Stolte2016}, \citen{Stolte2018}].
In 2017 NEXCO, one of the operators of expressways and toll roads in Japan, presented an automated snowplow vehicle.
A recent video demonstrates the current state of technology in which the GPS-guided vehicle autonomously removes snow from the road \cite{Driveplaza2021-vs}.

As one of the main road maintenance activities, the process of road-marking has also received a certain level of attention. 
In 2016, the company STiM from Belarus introduced  self-driving road-marking machine, `Kontur 700TPK'. In a video \cite{STiM_Company2016-cw}, they demonstrated a process of driverless paint application in which the vehicle presumably followed a given route using GPS-tracking. However, a lack of information makes it difficult to assess levels of system maturity and adoption in the industry. In 2018 Danish company TinyMobileRobots introduced a small GPS-guided robotic vehicle that can paint pre-markings\footnote{Pre-markings are used as guides on freshly paved roads to assist in the road-marking process.} onto the road.
However, this vehicle can only be used to assist the overall painting process because it cannot (re-)paint the road marking itself \cite{TinyMobileRobots2019-ai}.
Apart from a general motivation, the need for such driverless road-marking machines comes also from regulations such as `Work Zone Traffic Control Guidelines for Maintenance Operations' \cite{Wsdot2021-rl} or ASR 5.2\cite{Federal_Institute_for_Occupational_Safety_and_Health_BAuA2022-al} that specify a certain lateral spacing between a manned road-marking vehicle and the traffic to ensure safety.
In practice, this means that narrow construction sites require full closure to meet requirements given by the specification, which ultimately leads to inefficient traffic flow and increased costs.

To date, road-marking machines have not been officially used in a driverless fashion on public roads - especially not on roads without permanent separation from the traffic. This motivated the German Central Innovation Program (ZIM) to support the Project `System for Teleoperated Road-marking' (SToRM). The main goal of the project was to develop a driverless road-marking machine capable of performing road-marking maintenance tasks that satisfies both performance and safety standards.

\subsection{Contributions}
There are three main goals of driverless maintenance: increased safety of all road participants, more efficient traffic management, and reduced road maintenance costs that comply with the infrastructure level required for the adoption of \ac{ad}-powered mobility.
The aim of this paper is to give a perspective on how the current state of technology could be expanded to achieve those goals.
The paper presents the driverless road-marking machine from the system perspective considering both hardware and software elements as well as integration within the end-user ecosystem.
Presented operation modes are based on teleoperation and have been validated on the experimental vehicle developed within the scope of the `SToRM' project that provides a baseline for further efforts in this and similar areas.

%% file: tables/1_stakeholders.tex
\begin{table}[b]
	\caption{Estimated benefits for the major stakeholders through utilization of the driverless road maintenance technology}
	\label{tab:stakeholders}
	\centering
	\begin{tblr}{
			stretch=1.25,
			colspec={|Q[r,2.25cm]|Q[j,4.75cm]|},
			rowspec={Q[m]Q[m]Q[m]Q[m]}}
		\hline[1pt]
		\textbf{Stakeholder}  & \textbf{Benefit}\\
		\hline[1pt]
		Road Maintenance Crew & Improved safety and efficiency of maintenance. \\
		\hline
		Road Traffic  & Improved safety and traffic flow (time efficiency). \\
		\hline
		Road Infrastructure Authority  & Improved safety, traffic management, and infrastructure readiness for adoption of \ac{ad} technologies, while reducing relative maintenance costs. \\
		\hline[1pt]
	\end{tblr}
\end{table}

%% file: content/2_system.tex
\section{System Overview}

Compared to complex scenarios for highly-automated passenger vehicles such as urban environments, operating scenarios for road-marking machines have relatively simple constraints.
The vehicle operates in an isolated and restricted area, nominally without either static or dynamic objects it could collide with.
However, the vehicle still has to be able to perceive the environment and have suitable safety measures in place for unmanned operation on public roads.
These safety concepts have to account for possible system failures (both hardware and software), unwanted vehicle behavior such as lane/road departure, and unanticipated events caused by third parties. 
The driving trajectory is defined by road topology and can be scaled from quasi-straight roads with fast curves (e.g., highways), country roads with slower curves, to urban roads with slow curves and turns at different angles.
Additionally, vehicle dynamics are also considered to be trivial as the vehicle operates with velocities of no more than 7 m/s.
On the other hand, road-marking machines are equipped with a paint application module and need to perform road-marking tasks, which introduces a specific added complexity.
Table \ref{tab:scenarios} lists the three main operational scenarios:
\input{tables/2_scenarios.tex}

Different parameters and procedures have to be prepared and monitored during the application process such as sprayed line quality (positioning, length, width, thickness, etc.), reflective bead application (volume, spread quality, etc.), changing paint, refilling paint tanks, etc.
This complexity means that even if the vehicle can mark roads autonomously on its own, a certain level of expert remote assistance and/or remote monitoring will always be required.
For these reasons teleoperation has emerged as a technology that can introduce driverless operation mode to road-marking vehicles and help scale the level of their autonomy.

\subsection{Functional Description}
\label{subsec:opmodes}

A typical operational scenario for a driverless road-marking machine appears as follows: machine is transferred to a construction site.
Having arrived at the location, both vehicle and application systems are checked and prepared for the operation.
Once ready, the operator drives it remotely to the location where the road is to be marked.
The road-marking process can now start.
Depending on the mode of operation, a vehicle accelerates to reach the operating velocity either independently, or while being remotely driven by the operator.
At this point, paint application will start.
Once finished, the vehicle comes to a standstill and is returned to the safe zone under the remote control of the operator where a new mission can be prepared.

\subsection{Operation Modes and adopted Taxonomy}
\begin{figure}[!h]
	\centering
	\includegraphics[width=0.39\textwidth]{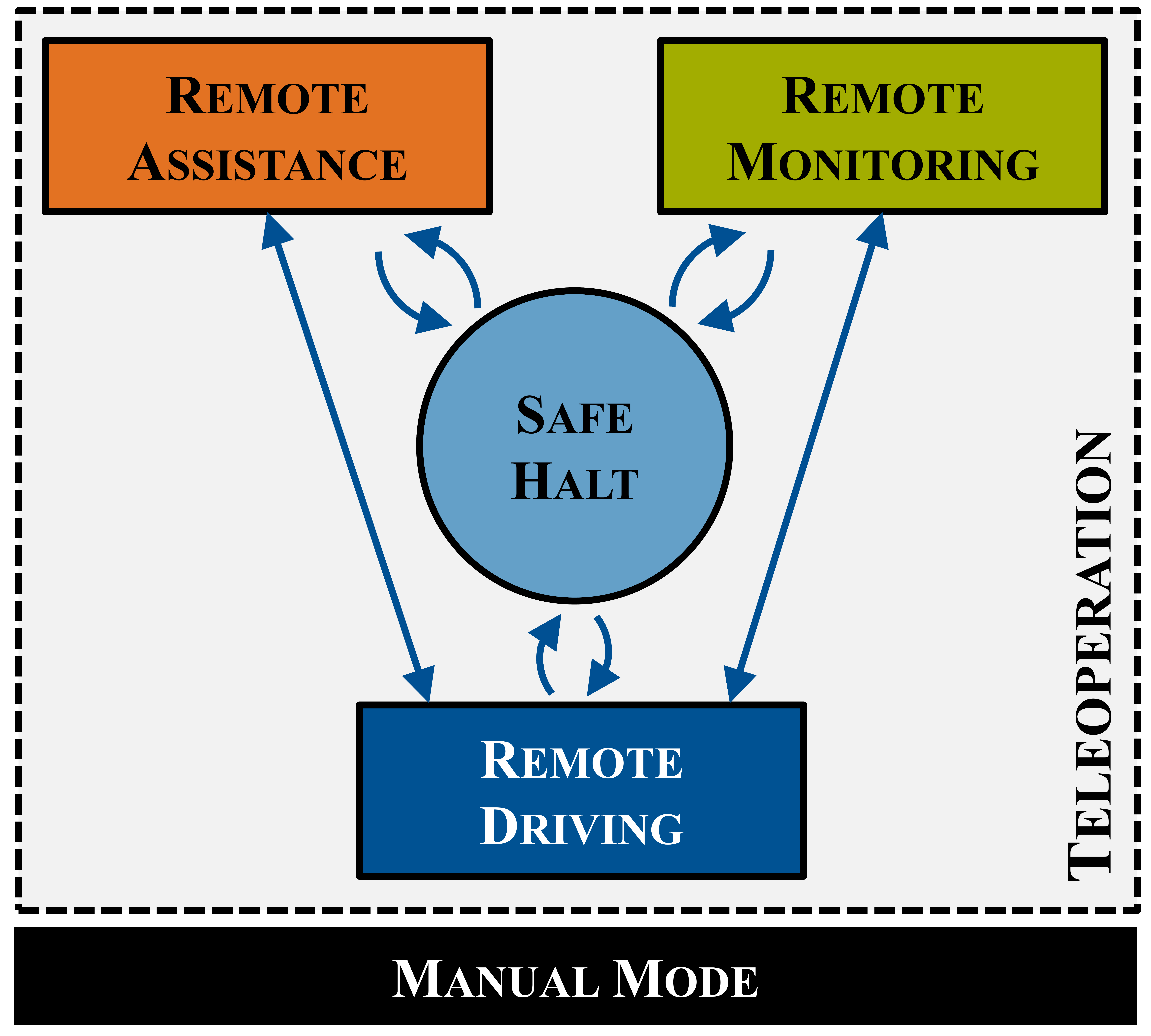}
	\caption{Operation modes of a driverless road-marking machine. State transitions based on the work of \cite{Stolte2015-kh}.}
	\label{fig:opmodes}
\end{figure}

Following the most recent classification of teleoperation interaction concepts [\citen{SAE2021}, \citen{CentreforConnected&AutomatedVehicles2022}, \citen{CentreforConnectedandAutonomousVehicles2020}, \citen{bogdoll}], five modes of operation have been identified and adopted as a base for operating driverless road-marking machines.
Corresponding state transitions based on the work of Stolte et al.\cite{Stolte2015-kh} are shown in Fig. \ref*{fig:opmodes}.

\vspace*{-4mm}

\textit{Manual Mode} allows the vehicle to be manually driven by an on-board human driver.
However, as the (driverless) technology matures, this mode, together with its hardware interfaces (steering wheel, pedals, etc.), will eventually be removed from the vehicle because it will no longer be necessary.
In \textit{Remote Driving Mode}, the remote operator has full control over the vehicle and its paint application system.
This mode, as well as all the other remote operation modes, has specific system requirements, interfaces and features, both hardware, and software in order to successfully fulfill the expected driving and paint application objectives.
\textit{Remote Assistance Mode} is a semi-automated mode in which a vehicle is capable of carrying out certain driving and/or paint application tasks autonomously, while being actively supported by a remote human operator.
This remote support can include a clear division of tasks between the operator and machine, or it might be realized in a shared control fashion [\citen{Anderson2013}, \citen{Schimpe2020steerwithme}].
It is anticipated that there is no one single configuration that will be ideal for use in all operational scenarios.
Instead, there might be a multitude of different configurations depending on road topology, road-marking type, machine/application system performance, or even the operator's personal choice.
From the vehicle automation point of view, \textit{Remote Monitoring Mode} represents the ultimate mode in which the vehicle has full control and can perform both driving and application objectives simultaneously.
The role of the operator is now completely decoupled and relaxed to the point where he/she merely oversees the work progress with limited intervention options.
Finally, \textit{Safe Halt Mode} incorporates a set of procedures that ensure the safety of the system. A vehicle can transition into this mode from any of the previously mentioned modes, either through operator intervention or by detecting violations of functional system boundaries and/or technical failures in the system. Table \ref{tab:taxonomy} provides an overview of the adopted taxonomy.

\begin{figure*}[!htbp]
	\centering
	\includegraphics[width=\textwidth]{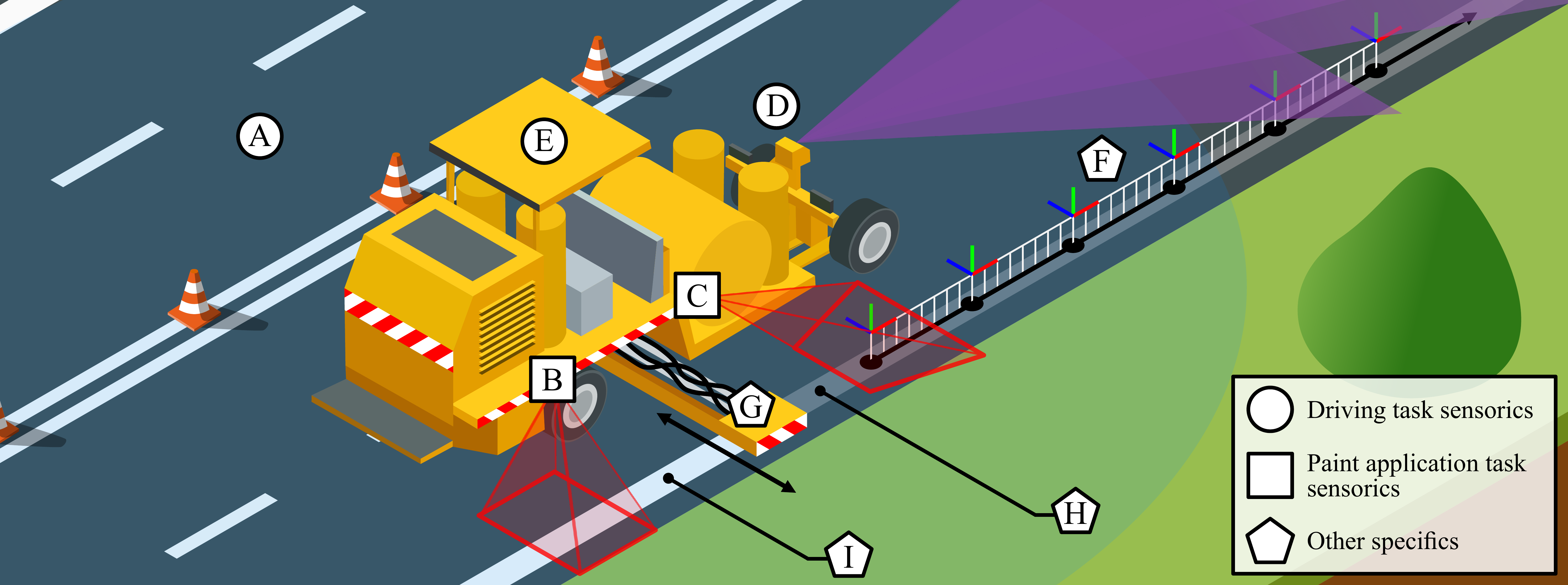}
	\caption{Proposed sensor setup to utilize driverless technology on conventional road-marking machines}
	\label{fig:machine}
	\vspace*{-3mm}
\end{figure*}

\begin{table}[!htpb]
	\caption{Adopted Taxonomy}
	\label{tab:taxonomy}
	\centering
	\begin{tblr}{
			stretch=1.25,
			colspec={|Q[c,0.5cm]Q[r,1.75cm]|Q[j,4.5cm]|},
			rowspec={Q[m]Q[m]Q[m]Q[m]Q[m]}, vline{2,2} = {2-4}{solid}}
		\hline[1pt]
		& \textbf{Term}  & \textbf{Description}\\
		\hline[1pt]
		\SetCell[r=3,c=1]{c} \rot{\textsc{Teleoperation}} & \textsc{Remote Monitoring} 
		& Remote operator monitors the system with limited intervention options. High level of vehicle autonomy.\\
		\hline
		& \textsc{Remote Assistance}  & Vehicle receives active remote assistance from the operator, while still being responsible for some parts of the driving/application objectives. \\
		\hline
		& \textsc{Remote Driving}  & Vehicle is under full remote control. \\
		\hline
		\SetCell[r=1,c=2]{r} \textsc{Manual Mode} & & Operator is on-board of the vehicle and manually operates it. \\ 
		\hline[1pt]
		
	\end{tblr}
\vspace*{-3mm}
\end{table}

\subsection{Road-marking Vehicle and its Sensor Stack}
There are now a large number of different teleoperated and/or automated vehicle types that have been introduced by the industry or are the result of research by academic institutions.
However, road-marking vehicles of the type discussed in this paper have received a very limited to non-existent coverage as described in Section \ref{subsec:SoT}.
Based on workshops carried out within the scope of project `SToRM', below, we present a proposal detailing the hardware and software required to overcome this shortcoming.

Road-marking vehicles are usually realized in two different fashions: either as a lorry that has been retrofitted with the paint application system or as a fully-fledged road-marking vehicle with its signature look as shown in Fig.~\ref{fig:machine}, both types having their own pros and cons mostly related to paint application specifics and transportation convenience.
Sensors required for driverless road-marking can be divided into two main groups: First group of sensors enables the driving task (both teleoperated and automated) and safety aspects, while the second group supports the paint application process and its \ac{qa}.
Fig. \ref{fig:machine} illustrates one such vehicle with marked positions of different sensors.
Positions designated with circles represent the first group of sensors, while the square-shaped marks denote the second group.
Position \krug{A} denotes a 360$^{\circ}$ visual coverage of the environment based on a multi-camera setup.
Position \krug{D} denotes a LiDAR sensor that scans the front corridor checking for obstacles.
Additionally, such a system could use a set of ultrasound sensors to improve the navigation on narrow sites.
Finally, position \krug{E} denotes the real-time kinematics sensor (GPS-RTK) used for global localization.
As part of the second sensor group, position \kvadrat{B} reveals the camera with its corresponding field of view.
This camera is positioned perpendicular to the road, and monitors the freshly painted road mark.
It provides a video feed to \ac{qa} software modules that assess the line quality against required standards and norms.
The second sensor \kvadrat{C} in this group is a pitched forward-facing camera which streams a video feed into spray gun control modules that estimate the offset between positions of the spray gun and the tracking line.
Finally, pentagon-shaped positions display some specific points of interest, for example, position \peterokut{G} denotes the spray gun of the paint application system, while positions \peterokut{H} and \peterokut{I} denote a road before and after the process of paint spraying.

\subsection{Software Modules}
\begin{figure}[!h]
	\vspace*{-5mm}
	\centering
	\includegraphics[width=0.45\textwidth]{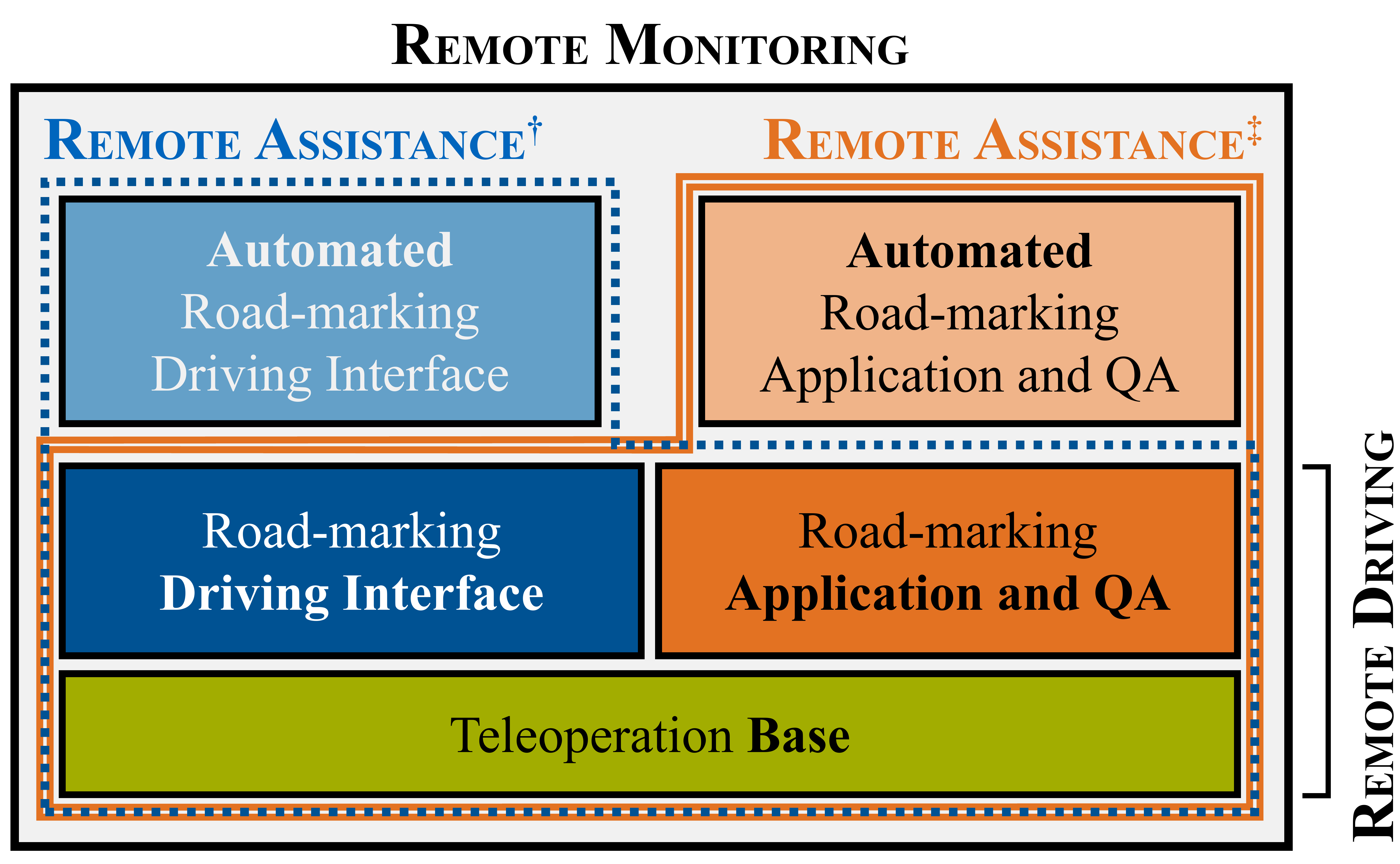}
	\caption{Software modules needed to utilize identified Operation Modes}
	\label{fig:sw_assistance}
\end{figure}

Driverless road vehicles are usually developed from the perspective of self-driving ability utilizing the well-known \textit{sense-plan-act} paradigm [\citen{handbook}, \citen{acm}].
This means that the vehicle has to perceive the environment, localize itself, detect and track other road participants, perform decision-making at all levels, and execute decisions.
Teleoperation generally complements this architecture by playing a smaller part that deals with unforeseen events and edge cases.
However, there are a number of reasons why teleoperation can be considered of major importance \textit{vis-a-vis} driverless road-marking machines.
First, the current level of maturity of this technology is very low. Initial use of teleoperation as a means for gaining experience and gradually implementing the \textit{sense-plan-act} features might be a reasonable way of scaling the level of autonomy with reduced overheads and in a much more efficient manner.
Second, the process of road marking, while a seemingly trivial task, has a lot of fine details that make it difficult to immediately jump to a high level of automation.
Use of skilled road-marking operators to teleoperate such machines is an excellent way of gaining insights about relevant working scenarios, gathering the learning data for the development of different software features, and exploring other initially unknown but important aspects.
There follows an introduction of high-level software components and features as depicted in Fig. \ref{fig:sw_assistance} which are needed to use the previously identified \textit{Operation Modes}.

\subsubsection*{1) \textit{\fauxsc{Remote Driving}}}
This mode is suitable for marking both old roads, and new roads with existing pre-markings (see Table \ref{tab:scenarios}).
At its core, the system is based on the teleoperation module named `\textit{Teleoperation Base}' which brings in all necessary features to establish the link with the remote vehicle, receive and display sensor data, and send operator commands.
The `\textit{Road-marking Driving Interface}' extends the core module with the low-level vehicle communication.
Additionally, it improves the operator's interface with features such as lane/line projections which guide the operator effectively during teleoperation in \ac{ar} fashion.
Finally, the `\textit{Road-marking Application and \ac{qa}}' module introduces vision-based line quality assessment that can be both displayed and stored as measurement results.

\subsubsection*{2) \textit{\fauxsc{Remote Assistance}}}
This mode features semi-automated driving/paint application where both remote operator and the vehicle work together.
Different task divisions are possible with various software feature flavors, but in essence either the remote operator will carry out the application task with the driving part being performed by the vehicle (RA$^\dagger$) or \textit{vice versa} (RA$^\ddagger$).
The `\textit{Automated Road-marking Driving Interface}' introduces functionalities that allow the vehicle to follow the specified path on its own.
Such a trajectory as depicted in Fig. \ref{fig:machine} with a position \peterokut{F}, can be produced either by use of specified plain GPS points, or vision-based detection of old markings.
The latter has many benefits because it can complement other functionalities and extend the operating domain in GPS-denied areas such as tunnels, etc.
The `\textit{Automated Road-marking Application and \ac{qa}}' relates to automated paint application. The system is capable of measuring the exact length of sprayed markings. The feed from camera \kvadrat{C} is used to estimate the offset between the tracking line and spray gun to minimize the overall path tracking error. Additionally, the system will automatically adjust the vertical position of the spraying head to ensure the newly sprayed line width remains within standard limits.

\subsubsection*{3) \textit{\fauxsc{Remote Monitoring}}}
The vehicle is capable of carrying out all the previously introduced features (on its own), thus effectively covering the full working envelope, while the operator merely monitors progress and has limited intervention options.

%% file: tables/2_scenarios.tex
\vspace*{-4mm}
\begin{table}[h]
	\caption{Main Road-marking operational Scenarios}
	\label{tab:scenarios}
	\centering
	\begin{tblr}{
			stretch=1.25,
			colspec={|Q[r,3cm]|Q[j,4.25cm]|},
			rowspec={Q[m]Q[m]Q[m]Q[m]}}
		\hline[1pt]
		\textbf{Scenario}  & \textbf{Description}\\
		\hline[1pt]
		Repainting old markings & Existing old markings are used as guidelines. \\
		\hline
		\SetCell[r=2,c=1]{c} Painting new markings  & (a) The road has been pre-marked with guidelines that assist the painting process.\\
		\hline
		& (b) The system uses GPS-data for guidance.\\
		\hline[1pt]
	\end{tblr}
\end{table}

%% file: content/3_concept.tex
\section{Proof of Concept}
Project `SToRM' offered a framework to develop and experimentally verify the proposed concept of a driverless road-marking machine system.
The following section gives details on implemented functionalities and features, as well as, overall system performance.

\subsection{Vehicle and Sensor Stack}
\begin{figure}[!htpb]
	\centering
	\subfloat[Experimental passenger vehicle next to improvised road markings during development phase testing various software features]{%
		\includegraphics[clip,width=0.95\columnwidth]{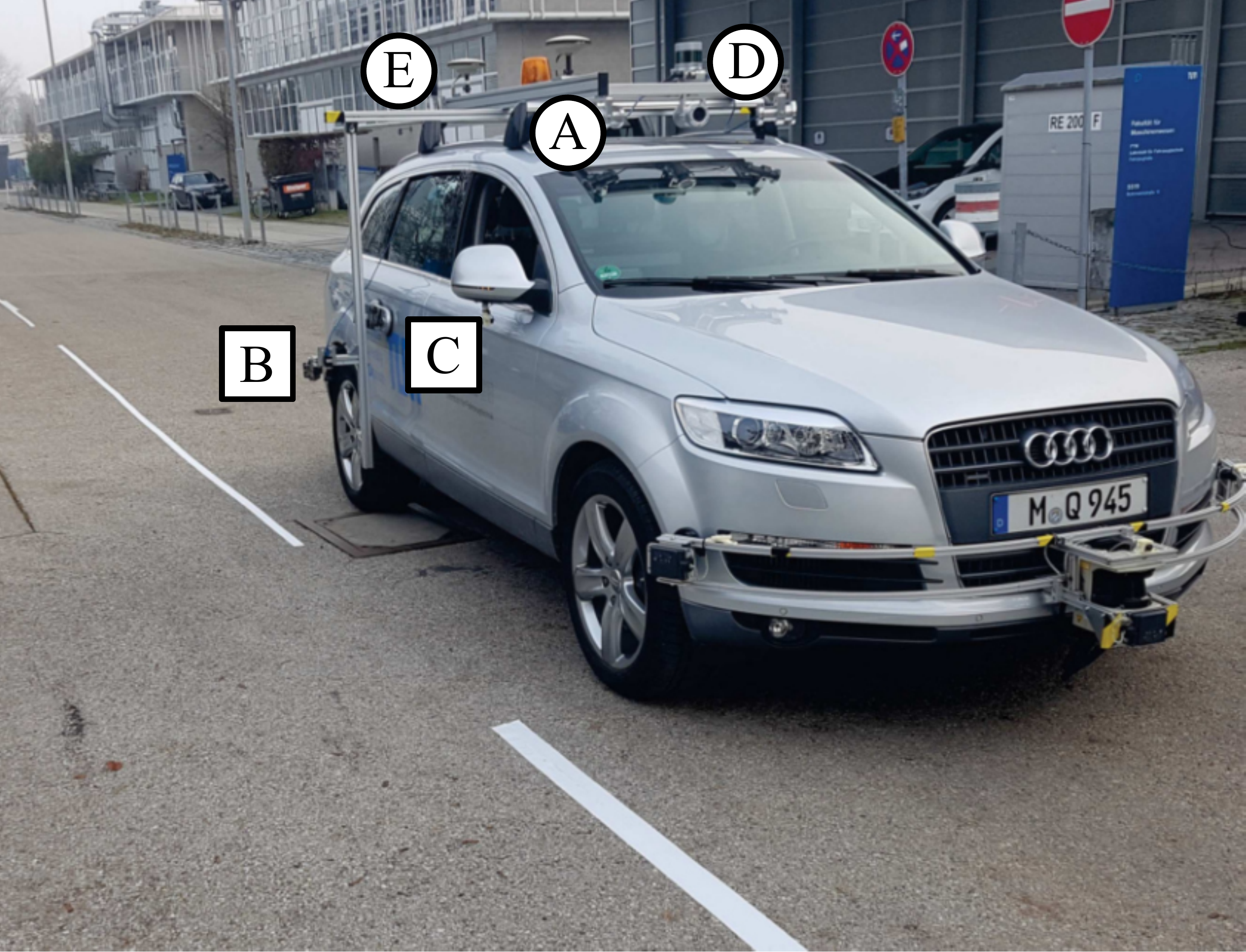}%
		\label{fig:storm_1a}}
	\\
	\subfloat[Experimental road-marking machine during teleoperation on the premises of the Grün company in Siegen, Germany]{%
		\includegraphics[clip,width=0.95\columnwidth]{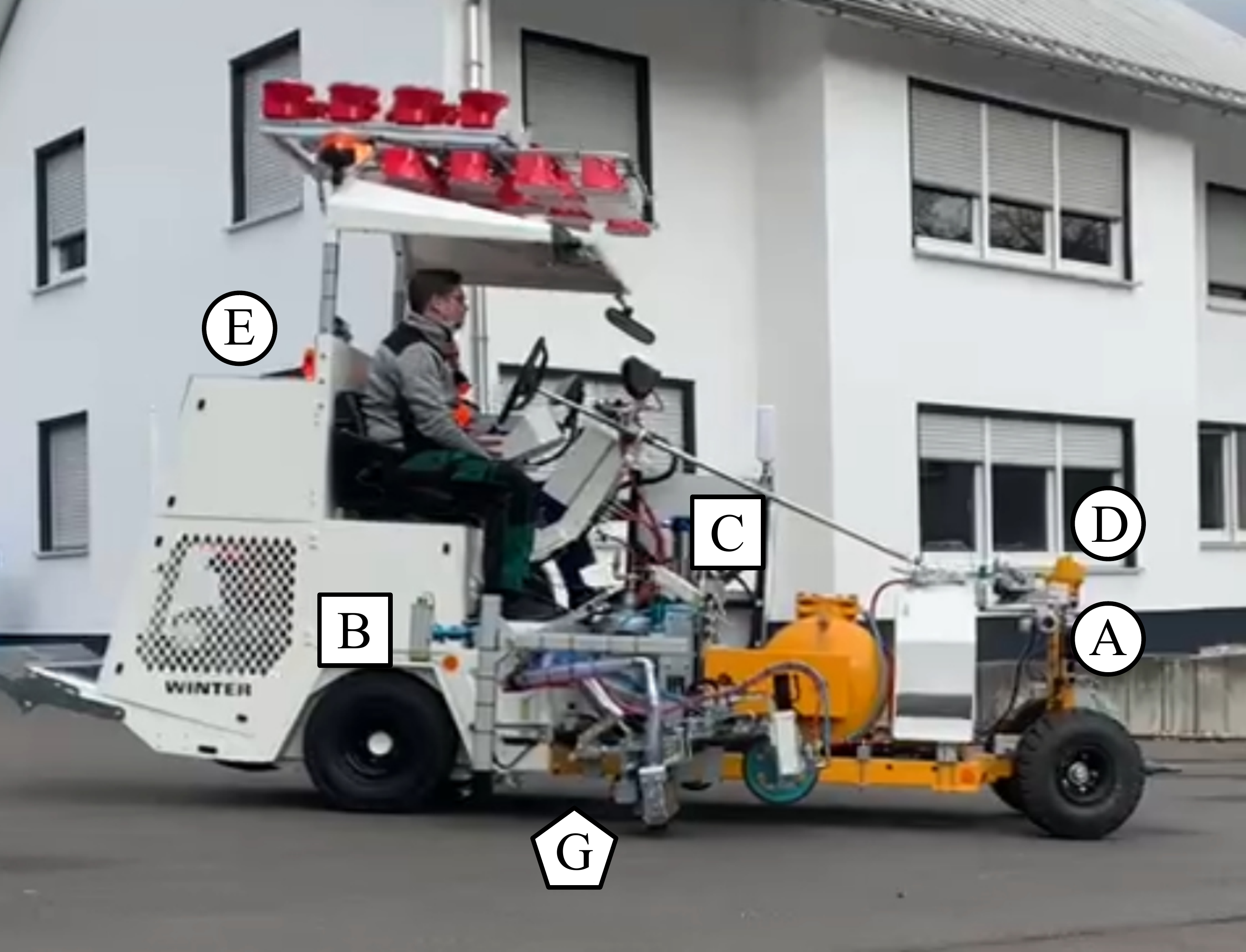}%
		\label{fig:storm_1b}}
	\caption{Vehicles used for development and experimental verification of the presented driverless road-marking vehicle concept}
	\label{fig:storm_1}
\end{figure}

Fig. \ref{fig:storm_1} shows two experimental vehicles used as prototypes.
Fig. \ref{fig:storm_1a} shows a passenger car and Fig. \ref{fig:storm_1b} shows a current road-marking machine. Both systems were equipped with \ac{cots} sensors in a manner similar to that portrayed in Fig. \ref{fig:machine}. Additionally, Table \ref{tab:storm_sensors} gives a sensor overview and indicates some minor setup differences between the vehicles.

\begin{table}[t]
	\caption{Experimental Vehicles - Hardware Overview}
	\label{tab:storm_sensors}
	\centering
	\begin{tblr}{stretch=1.5, colspec={Q[c,1.5cm]Q[c,2.5cm]Q[c,3cm]},rowspec={Q[m]Q[m]|Q[m]|Q[m]|Q[m]|Q[m]|Q[m]},row{1} = {bg=whitesmoke}}
		\hline[1pt]
		\textbf{Position}  & \textbf{Description} & \textbf{Product} \\
		\hline[1pt]
		\krug{A} & Vehicle Control Camera & FLIR BFS U3-28S5M-C \& EO 5mm f/2.8 FL \\
		\kvadrat{B} & Quality Assessment Camera & FLIR BFS U3-28S5M-C \& EO 16mm f/1.8 HPr \\
		\kvadrat{C} & Spray Gun Control Camera & FLIR BFS U3-28S5M-C \& EO 16mm f/1.8 HPr \\
		\krug{D} & LiDAR & {Velodyne Puck\textsuperscript{*}\\ Livox Mid-40\textsuperscript{$\dagger$}} \\
		\krug{E} & GPS-RTK System & {OXTS RT3000 v3\textsuperscript{*}\\ Emlid RS+\textsuperscript{$\dagger$}} \\
		\peterokut{G} & Spray Gun & {N/A\textsuperscript{*}\\ Winter Markiertechnik\textsuperscript{$\dagger$}} \\
		\hline[1pt]
		\SetCell[r=1,c=3]{l} \textsuperscript{*} Passenger vehicle, \, \textsuperscript{$\dagger$} Road-marking vehicle
	\end{tblr}
\vspace*{-5mm}
\end{table}

\subsection{Software Modules}
\subsubsection{Teleoperation Base}
\label{subsec:tofbase}

Conceptually inspired by the system design introduced by \cite{Gnatzig2013-xg}, this module utilizes a highly modular open-source software stack for teleoperated driving based on \ac{ros} middleware developed and maintained by Schimpe et al. \cite{Schimpe2022-ig}.

\subsubsection{Road-marking Driving Interface}
\begin{figure}[!h]
	\centering
	\includegraphics[width=0.47\textwidth]{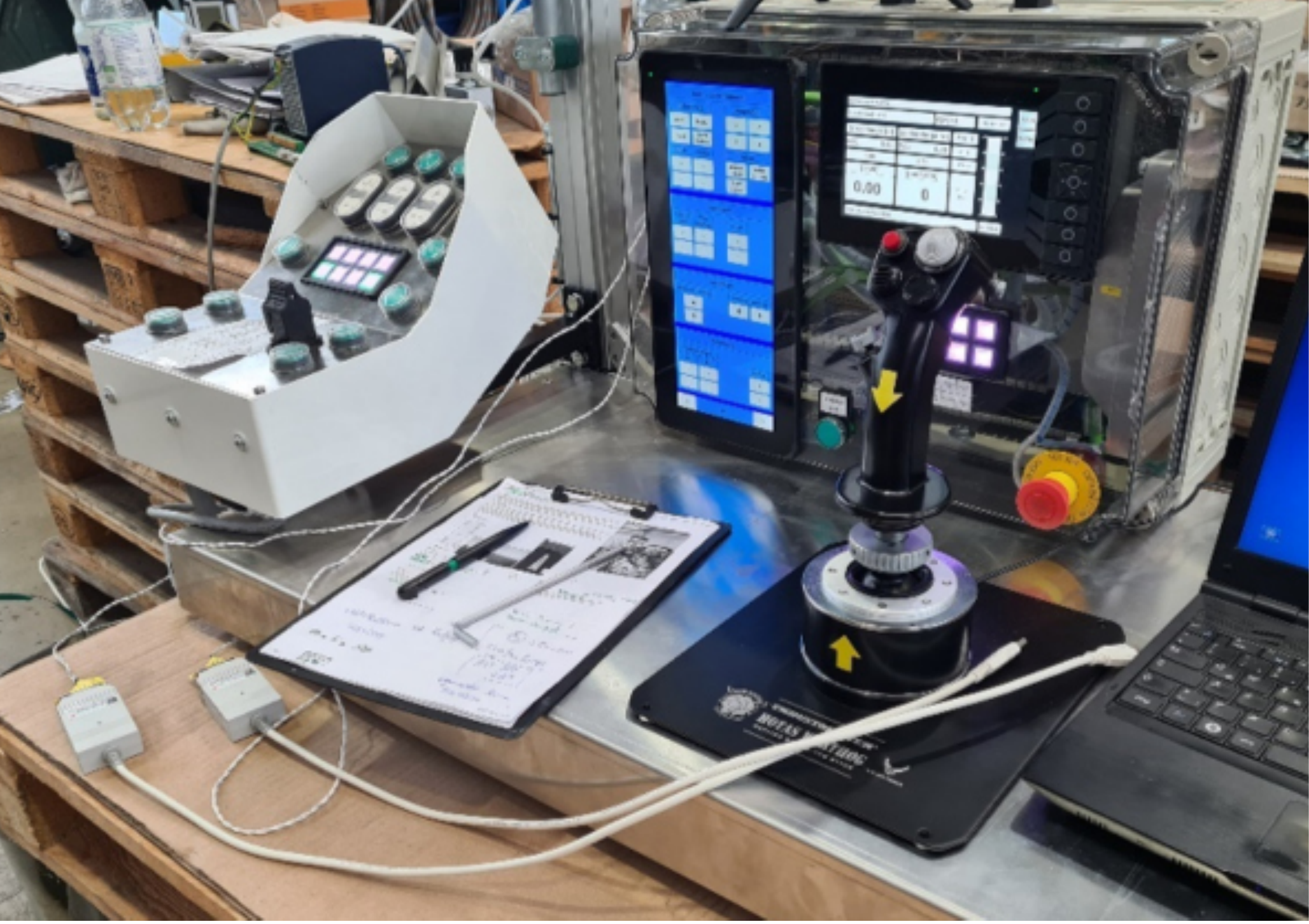}
	\caption{Concept of the control center for teleoperation of road-marking machines}
	\label{fig:storm_6}
\end{figure}

\begin{figure}[!h]
	\centering
	\includegraphics[width=0.47\textwidth]{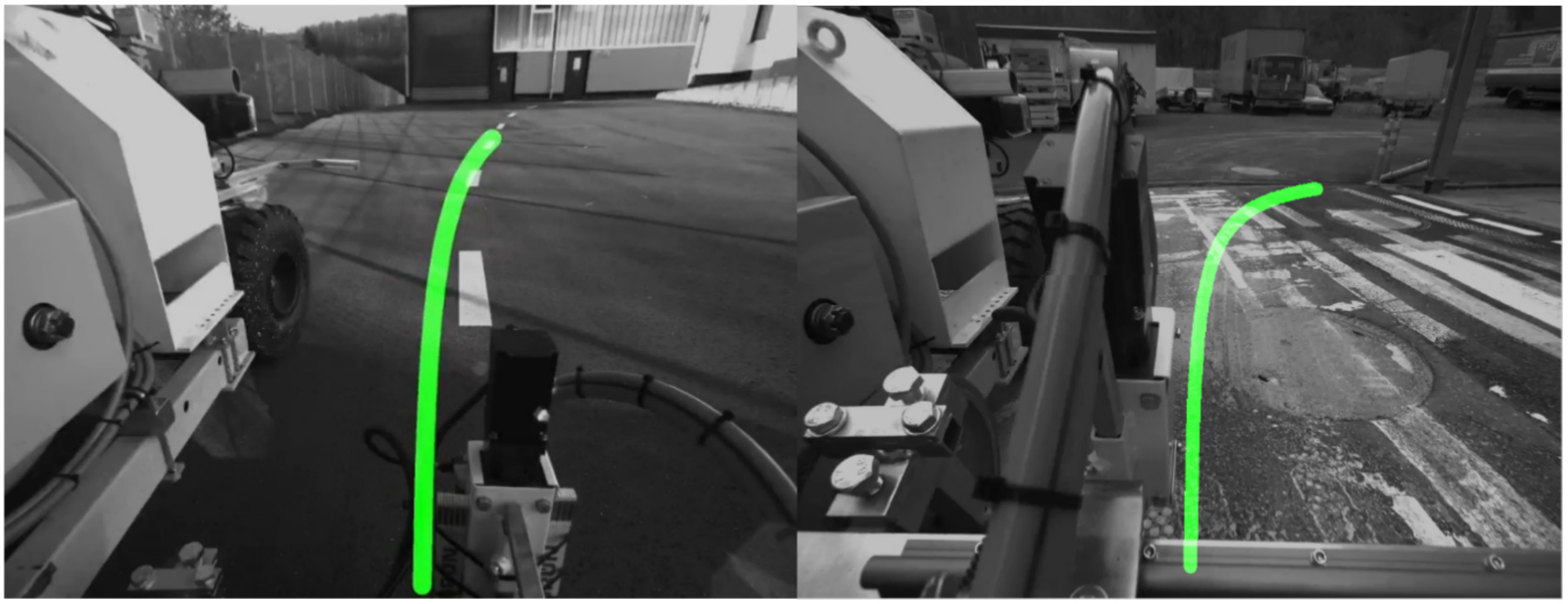}
	\caption{Steering wheel angle-based projections of the spraying course in two separate scenarios}
	\label{fig:storm_2}
\end{figure}

The software base in subsection \ref{subsec:tofbase} has been extended to create a communication interface with the road-marking vehicle shown in Fig. \ref{fig:storm_1b} with the hardware interface being realized using two CAN-BUS networks. Fig. \ref{fig:storm_6} indicates how the control center was conceptualized. The human operator uses a joystick, and a combination of touchscreens and panels with physical buttons to interact with the vehicle, while observing the system and sensor feeds via a notebook computer. Additionally, the projection module of the ToD Software \cite{Schimpe2022-ig} has been adapted to project the spraying line and precisely guide the remote operator during teleoperation process as shown in Fig. \ref{fig:storm_2}.

\vspace{-5mm}

\subsubsection{Road-marking Application and \ac{qa}}
\begin{figure}[!h]
	\vspace*{-2mm}
	\centering
	\subfloat[Assessment of the spraying performance on a 15-second detection sample. A frame showcasing the width estimation alogrithm (left). Estimated line width per frame (right).]{%
		\includegraphics[clip,width=0.95\columnwidth]{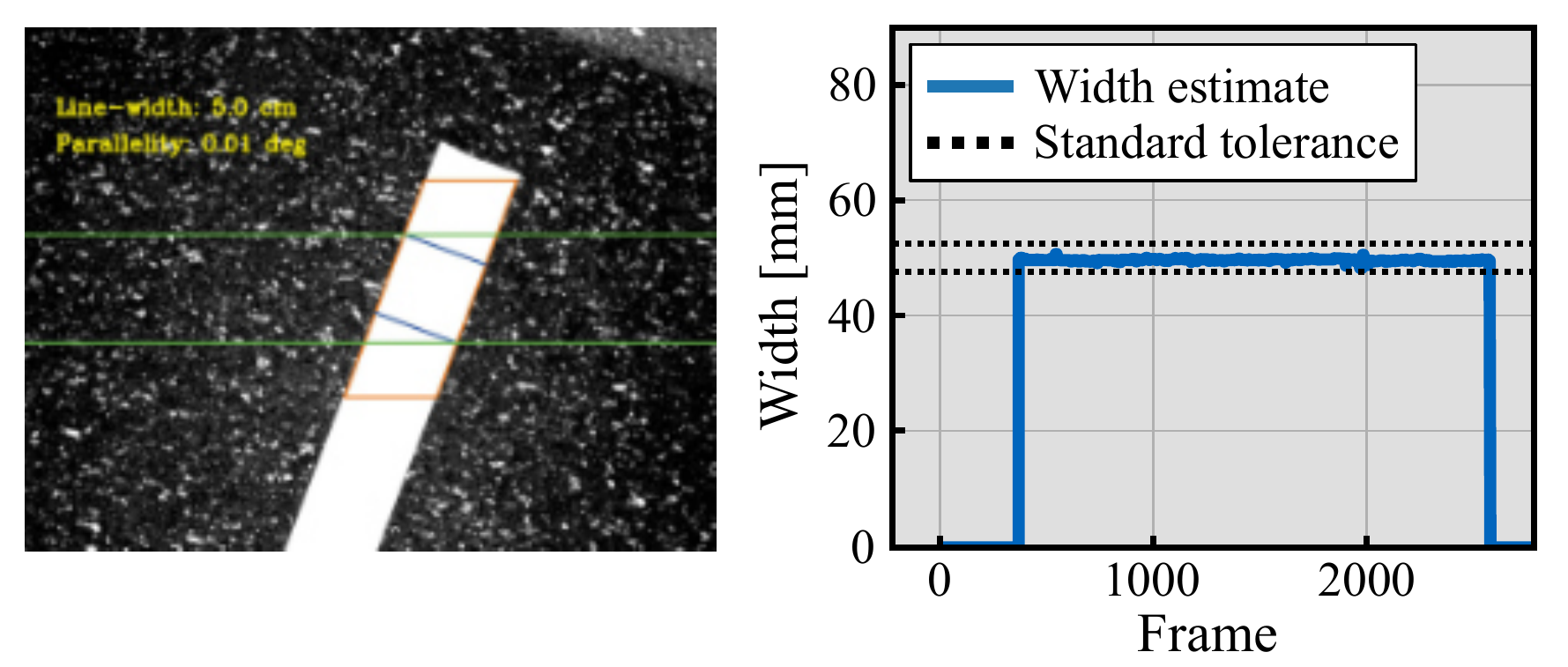}%
		\label{fig:storm_7a}
	}
	\\
	\subfloat[Vision-based velocity estimation (green) compared with the ground-truth velocity (black) in the upper graph, and the corresponding estimation error (blue) shown in the graph below. ]{%
		\includegraphics[clip,width=0.95\columnwidth]{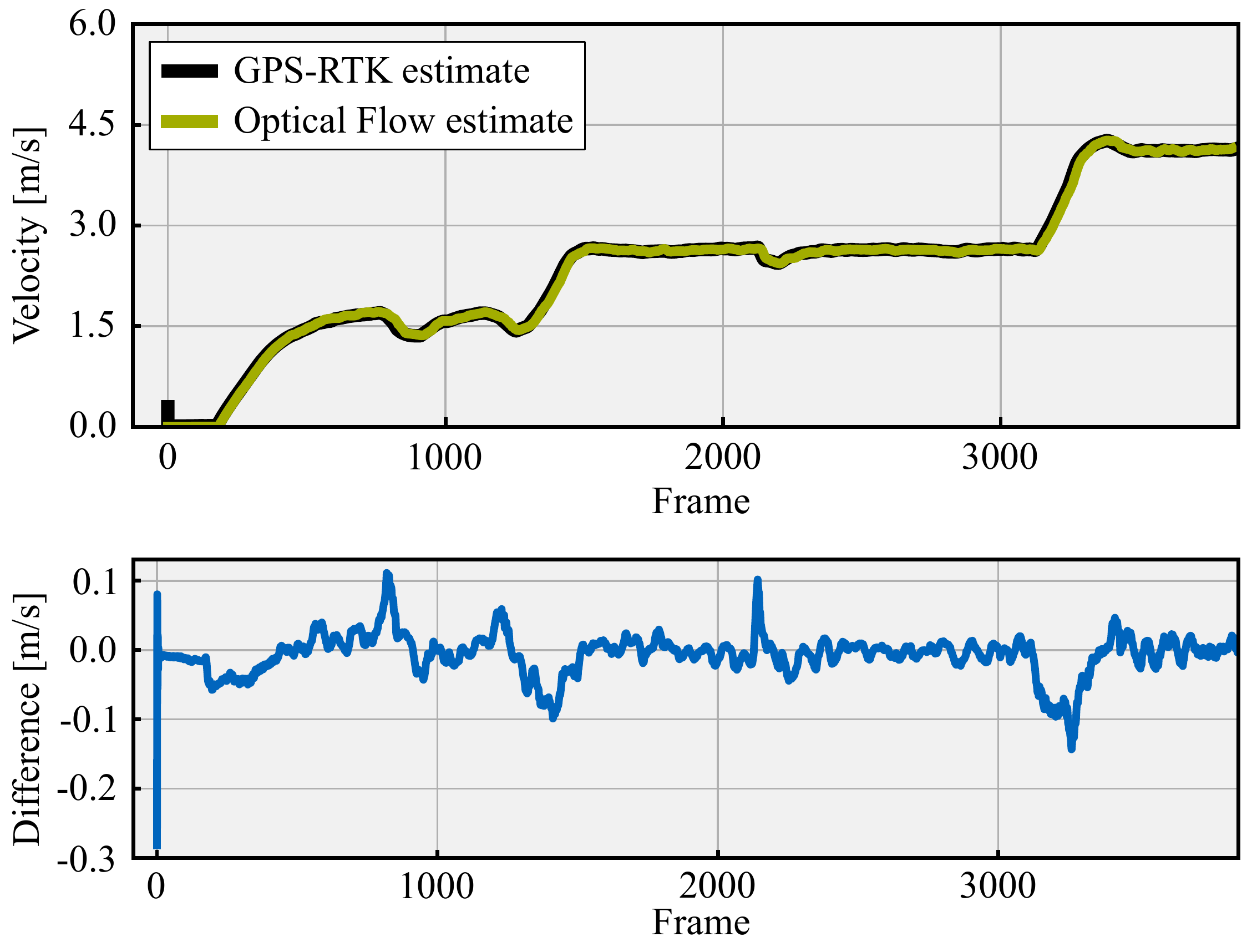}%
		\label{fig:storm_7b}
	}
	\caption{Performance of the `\textit{Road-marking Application and \ac{qa}}' module}
	\label{fig:storm_7}
\end{figure}

This module has been developed and integrated to help the operator assess the spraying performance.
It receives the image stream from the camera \kvadrat{B} and analyses the image, while effectively measuring the line width.
This information, together with the snapshot of the road-mark, is then transferred to the control center where the remote operator can evaluate the results.
Fig. \ref{fig:storm_7a} shows one frame produced by the module (left) and a corresponding 15-seconds long drive detection/estimation sample (right).
The module has a number of subroutines that ensure robust and reliable width estimation.
Given that the system is equipped with a (previously mentioned) downward-facing camera \kvadrat{B}, a vision-based velocity estimation module has been developed with the main aim being to provide an accurate velocity estimate.
In this way the vehicle can receive reliable velocity information even in GPS-denied areas such as tunnels. 
Here, a custom Lucas-Kanade-based optical flow algorithm was developed.
Coupled with filtering heuristics it proved to provide a reliable velocity estimation up to 8 m/s for a fraction of the price of advanced GPS systems.
Fig. \ref{fig:storm_7b} shows the performance of the velocity estimation compared to the baseline OXTS RT3000v3 GPS-RTK system.
However, because it is heavily dependent on good lighting conditions to keep the exposure time as low as possible to avoid motion-blur, a whole-day missions or activities during different seasons might require an external source of light.
Finally, a dataset with camera image streams has been published to support the future development of  algorithms for this and similar application(s) \cite{mediatum1685449}.

\subsubsection{Automated Road-marking Driving Interface}
\begin{figure}[!h]
	\centering
	\includegraphics[width=0.47\textwidth]{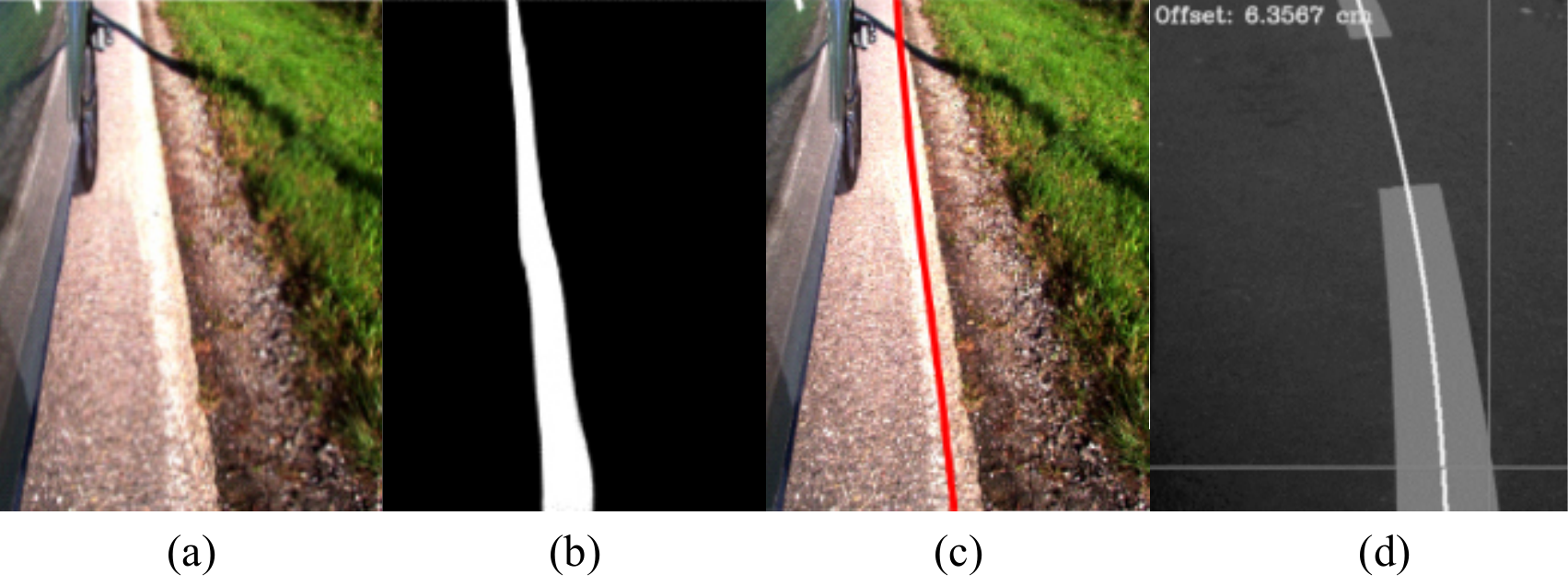}
	\caption{CNN-based detection of old road markings: Test sample (a), Inference (b), Fitted trajectory (c). Spraying gun (tracking) offset estimation~(d).}
	\label{fig:storm_4}
\end{figure}

This module introduces three major software features that make automated driving possible.
The first one is the deep neural network based detection of old markings.
The idea is to detect those markings and use them as a basis to construct reference (tracking) trajectories.
However, since old road markings are usually damaged and in a bad shape, such detection is not possible with conventional methods.
Instead, a \ac{cnn} was trained on a custom image dataset that was manually collected and labeled.
Once deployed, the system demonstrated robust and efficient trajetory estimation.
Fig. \ref{fig:storm_4} shows the three image samples, namely: dataset image (a), inference result (b), and an estimated path shown in red (c) that is then converted into a trajectory within the local coordinate system.
The second feature is support for GPS-based trajectories which makes the road marking process much more convenient and efficient because it does not require any preexisting marks or preparation procedures. 
Finally, the third feature is a tracking controller that generates the vehicle control signals to track the given trajectory.
Here a number of controller strategies have been deployed and evaluated within the kinematics bicycle model framework, such as PID, Stanley, Pure Pursuit, Adaptive Pure Pursuit, etc. 

\subsubsection{Automated Road-marking Application and \ac{qa}}
Once the vehicle can track a trajectory, it will always exert some level of tracking error.
To cope with this, the lateral position of the spraying gun can be controlled.
The anticipated tracking error is estimated using camera sensor \kvadrat{C} with the sample detection frame shown in Fig. \ref*{fig:storm_4}(d).
Tracking information is fed into the linear positioning system that will reposition the spray gun position accordingly.
Additionally, the system can control the vertical linear position of the spraying head, measure the length of both road markings and gaps, and also trigger the spraying system on its own, thus automating the whole process.\\

\subsubsection{Other Modules}
Besides the modules introduced so far, there are other system specifics that are essential for the successful operation of the vehicle.
The presented system has a number of safety features in place.
The \textit{Safe Halt} mode not only makes sure that the vehicle will reach a stand-still state safely, but will also trigger additional safety measures to flush the paint application system and keep it safe.
The system network supports both wireless and cellular networks with the latencies measured around 100 ms.
Additionally, the connection between vehicle and operator has been realized within a virtual private network (VPN) for added communication security.

\subsection{System Performance and Future Work}

An experimental verification took place on the premises of the Grün company in Siegen, Germany, with different locations of the remote control center - from being at the same test area in Siegen, up to a remote location approximately 500 kilometers away, in Munich.
Tests have included different driving scenarios, road marks, lighting and weather conditions, and were carried out on a reduced scale, since full-scale experimental verification on actual construction sites was beyond the scope of the project. Results have shown that the introduced remote operation concept can be used to teleoperate the road-marking machine in a way that fully meets the given requirements.
Implemented assistance systems have demonstrated that they are able to provide reliable and robust information also significantly improving upon the existing manual mode experience.
Having a joystick instead of a steering wheel for remote driving did not prove to bring significant benefits apart from requiring less space, however, this requires further verification with more realistic driving scenarios and longer operating sessions. Future work should scale up the test activities to actual construction sites and more relevant test scenarios.
This would also include fine-tuning of the existing modules, and introduction of new ones if needed to support new features such as shared lane markings, etc.
Additionally, studies should be designed and conducted to assess the level of situational awareness and mental load of the remote operator.
This would also be used to improve the overall control center design and further assess means of physical interaction with the system.
Finally, to bring the vehicle closer to a market introduction, a systematic safety assessment is required as well as consultations with legislators and other relevant authorities prior to the final introduction of these vehicles to public roads.

%% file: content/4_conclusion.tex
\section{Conclusion}
Driverless road maintenance could potentially be highly beneficial to all its stakeholders.
This paper has demonstrated how driverless technology might be deployed to road-marking vehicles. 
Taking teleoperation technology as the system base, the paper proposed a set of operating modes and functionalities that can scale the operation of such vehicles from plain remote driving to remote assistance and monitoring where the human operator assists or merely monitors the system.
The concept was developed and successfully tested with an actual experimental road-marking machine, creating a baseline for future efforts in this and similar areas.

%% file: content/acknowledgments.tex
\section*{Acknowledgements}
Domagoj Majstorovi\'c as the first author, was the main developer of the presented work. Frank Diermeyer made essential contributions to the conception of the research project and revised the paper critically for important intellectual content. He gave final approval for the version to be published and agrees to all aspects of the work. As a guarantor, he accepts responsibility for the overall integrity of the paper. The authors want to thank their project partners, companies Grün, GmbH and TeaX, GmbH for their support, as well as, students J. Bömelburg-Zacharias, M. Mija\v{c}evi\'c, M. Peri\'c and Y. Xue. The research was funded by the Central Innovation Program (ZIM) under grant No. ZF4648101MS8.